# Rapid Face Mask Detection and Person Identification Model based on Deep Neural Networks

**Abdullah Ahmad Khan**[1], **Mohd. Belal**[2], **and Ghufran Ullah**[3]
[1,2,3]Department of Computer Science, Aligarh Muslim University, Uttar Pradesh, India

[1]**aakhan3@myamu.ac.in**
[2]**mohdbelalalig@gmail.com**
[3]**ghufranullah1997@gmail.com**

## ABSTACT

As Covid-19 has been constantly getting mutated and in three or four months a new variant gets introduced to us and it comes with more deadly problems. The things that prevent us from getting Covid is getting vaccinated and wearing a face mask. In this paper, we have implemented a new Face Mask Detection and Person Recognition model named Insight face which is based on SoftMax loss classification algorithm Arc Face loss and names it as RFMPI-DNN(Rapid Face Detection and Peron Identification Model based on Deep Neural Networks) to detect face mask and person identity rapidly as compared to other models available. To compare our new model, we have used previous MobileNet_V2 model and face recognition module for effective comparison on the basis of time. The proposed model implemented in the system has outperformed the model compared in this paper in every aspect

## 1. Introduction

As the corona-virus case are increasing day by day although the vaccination drive has been going on at the full speed but still only 29% of total world population has been fully vaccinated yet. In India only 12% of people have been vaccinated and it assumed that next wave of newly mutated corona-virus is coming in mid-October, so getting vaccinated and wearing a face mask is still a requirement. As we have also seen the rise of pollution in different cities so wearing a face mask is also beneficial for one's health. By implementing our proposed project, we can avoid spread of Covid-19 which is necessary and beneficial for large institutions and businesses so that they can maintain their productivity while avoid getting infected to COVID-19 and also encourage people to wear mask while commuting so that they don't breathe polluted air and be safe. As the quality of air is getting worst day and in some cities the air is so bad that the government has suggested people to wear a face mask so that people can breathe safely and does not inhale air which is mixed with toxins. But since 2020 the world has been dealing up with the pandemic i.e., COVID-19. The earlier symptoms of it were having high fever and difficulty in breathing etc. and the people who had no previous medical records are also getting infected with the virus.

It is advised by every major Health Organizations that wearing face mask and getting vaccinated and washing hands with alcohol-based sanitizers. The first two part we need to be governed by ourselves, the proposed model will make aware people about the safety of wearing a face mask not only for avoiding polluted air but also for others safety. In this paper we have tried to minimize the time taken by detection model to detect and predict the person and his/her identity. The InsightFace module has been incorporated in place of previous module to reduce effective time taken during the computation. InsightFace is an incorporated python library for 2d & 3D face analysis. InsightFace effectively executes a rich variety of best in- class algorithms of face recognition, face identification and face alignment, which upgraded for both training and deployment. The contribution of this research work is as follow:

- A SoftMax loss-based module is incorporated i.e., InsigtFace for Detection and Prediction.
- The simulation is performed on Wider Face benchmark to check the efficiency of the proposed module.

The images we have used in this project for visual object recognition are gathered from ImageNet. ImageNet has a mega database of different ordinary images classification from having hundreds of images of "strawberry" to "inflatable. It is database which mainly focuses on visual object recognition. It contains about 20,000 classifications and about 14 million pictures. The presented project is measured on ImageNet. For getting prediction on masked and non-masked face and person identification we have used Deep Learning network. Deep Learning is part of Machine Learning which is neural network that has 3 more layers. These neural networks in the deep learning model help us to simulate the behaviour of human brain like learning huge amounts of data and matching the ability to the real human brain. In previous generations we have seen AI and Machine Leaning working on some amazing tasks like self-driving car, feasible working of websites etc. These things have become so common today we see them a lot in our real life but we ignore them.

The rest of the paper is divided into 7 sections given as Section 2 discusses Related Work on different modules of face detection and recognition. In Section 3, we presented the problem formation focusing in the existing system. Section 4 presents the proposed work, including the overview of ArcFace algorithm based on Soft Max loss. Section 5 gives Simulation study for benchmark on the proposed module. Section 6 presents the Analysis, Results. Finally, Section 7 talks about the conclusion and future work part of the paper.

## 2. Related Work

[1] this paper titled "Deep learning based safe social distancing and facemask detection in public areas for COVID-19 safety guidelines adherence in this research paper author have built a real-time system integrated with security cameras in the public places which detects whether the person is wearing a face mask or not and also check if heaa/she is maintaining proper social distance. The system reports directly to the authorities and it was built using raspberry pi and OpenCV.

[2] they have designed a hybrid deep learning model using two components. Resnet50 being the first component is used for feature extraction and for classification they have used decision tree and Support Vector Machine (SVM) algorithms. For simulation study they have used 3 datasets to benchmarking their model Simulated Mask Face Dataset (SMFD) is the first dataset Labeled Faces in the Wild (LFW) is the second dataset and Real-World Masked Faced Dataset (RMFD) is the third one. The results achieved LFW in the SVM algorithm was 100% for the RMFD it achieved 9.64% and for SMFD it gets 99.49% accuracy.

[3] In this paper a high accuracy and efficient detector called as "Retina Face" was developed on stage which consist of feature pyramid network to combine several feature maps. Transfer learning is applied to reduce the shortage of datasets. The approach of this model was to eliminate the projections of poor confidence and high Union Intersection. The accuracy result for the Retina Face was high developed on the dataset of face mask. It achieved 2.3% which was 1.5% higher than standard results and achieved detection precision of 1.0% which was 5.9% higher than the previous standard results.

The paper proposed by the author Toshanlal Meenapal [4] Put together couple of face classifier which identifies any face in any conditions. This paper proposed a novel technique uses that uses convolutional neural network model VGG-16 predefined weights etc. Convolution Networks are used to fragment the faces from the image/picture. For utilizing misfortune work Binomial Cross Entropy is used and for preparing they have used Angle Descent. FCN is used to eliminate undesirable noise and to avoid false expectations. Results of the model got the accuracy it got for the portion where the face mask is used to cover the face was 93.844%.

[5] In this paper previous built state-of-the-art deep learning model named InceptionV3 is fine tuned. For training the dataset the author has used SFMD, i.e., Simulated Face mask Dataset which is used to simulate current set of data on a given dataset which is available publicly for better testing and training of images. Image augmentation is used so that we can remove the data restriction. The proposed model attains 100% accuracy during the testing and 99.9% precision during training [6].

[7] proposed a model named Smart City and Intelligent Transport System which uses IoT devices and different sensors to enforce the social distancing so that people can follow government guidelines promptly. For monitoring real time movements of objects their model describes deploying sensors in different parts of the city and it also offers data sharing. Overall, the system architecture has the ability to store and share data and information to the relevant central cloud facilities and also has the ability to record the data at real time and exchange messages with nearby sensors.

[8] The authors in this paper have discussed about the contribution of Smart Cities in containing the spread of COVID-19. They have specifically talked about smart cities of South Korea. Not only they tracked people of wearing a mask in public and maintaining social distance but it they also tracked the user medical history, purchase history, cell phone

location, even the real time monitoring of people not only on the public places but even in their buildings which helps the government to suppress the COVID-19 infection in the South Korea much better than any other countries.

[9] Have talked about how IoT can help to suppress COVID-19 in healthcare sector and can save a lot of life. The developed system gave emphasis on interconnect devices so that hospital can keep track of different cases. They have given major key-merits for IoT that can help to fight COVID-19 pandemic such as when the system is operated by IoT device there will be lesser chances of errors as compared to work done by humans. IoT devices will provide effective control and will enhance the diagnosis. They have talked about several application of IoT in the healthcare sectors like Automated treatment process, Telehealth consultation, Wireless Healthcare network to identify COVID-19 patients, Rapid COVID- 19 screening, connecting all medical devices through the internet, accurate forecasting of virus etc.

[8] has talked about state-of-the art model which helps to suppress the COVID-19 pandemic without having a lockdown in the entire country. The paper titles "Smart city technologies for pandemic control without lockdown" in this research they interview different patients and their past movement has been recorded. They have noticed some of the patients have hidden their past records from the interviewer. But by using the Realtime tracking we can get the information accurately.

[10] has proposed a unique way to deal with the pandemic using the position of technology to track infected people. They proposed the use of Drones and Robot technologies as medical personnel to deal and track the infected patients.

[11] has used Principal Component Analysis (PCA) to detect and recognize face images. They have used small dataset for detection of faces in the images. They worked on the GUI of the model. The model has various buttons and text fields such as selecting an image, running the PCA recognizer to recognize the selected image from the image database. If the user is recognized then the selected image and the image from he database is shown side by side and text field below the image shows the name and grant access to the system and if the person is not in the database, then the text field will show access denied, until he user clicks on register them as a new user.

## 3. Existing system

The existing system deals with Mobile_NetV2 algorithm for classification and prediction. The system used 20% images for test and rest for 80% training purpose. The existing system uses face recognition library from python to recognize faces and identity of the person [12]. The system helps to identify the person wearing mask or not but it fails do it efficiently and sometimes returns ambiguous results.

**3.1 Issues in existing system**

The existing systems does not allow the system to easily scale and maintain real time inference capability when the number of users increases.

**3.2 Drawbacks in existing system he major limitations of existing schemes are as follows: -**

- Mobil_NetV2 algorithm is used to train and test the data in the existing which is slower as compared to new generation models.
- The existing system does not have the inference capability and does not use powerful GPU to compute the results.
- The facial recognition system used in the existing system is old and does not cover complete face.

## 4. Proposed system

In the proposed work we have incorporated a new Deep Learning Neural Network named as InsightFace in place of old face recognition library. The InsightFace used MXNET (DNN Framework) as its inference backend which allows our system flexibility, fast model training with high scalability which helps in maximize productivity and efficiency.

The InsightFace framework uses face detectors with ArcFace (Additive Angular Margin Loss for Deep Face Recognition) algorithm to obtain highly discriminative features for face recognition. The ArcFace uses modified SoftMax loss classification that makes predictions between the feature and weights.

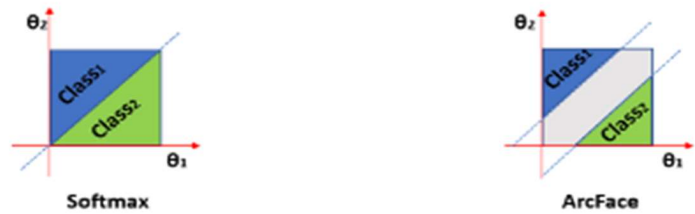

*Figure 1* Decision margins of different loss functions under binary classification case. The dashed line represents the decision boundary, and the grey areas are the decision margins.

The proposed model of face recognition implements singleton class which creates a single instance of an object which ensures our model will be time and memory efficient. By implementing the InsightFace bounding box wrapper we allow our model to detect the image positions as well as object of interest in the images and is represented by rectangular boxes, and with the help of embedded wrapper, it performs feature selection during the model training.

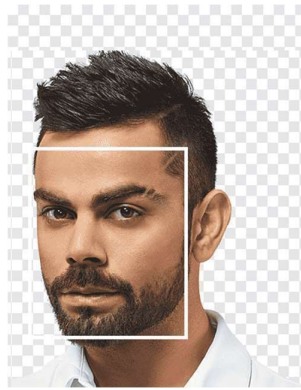

*Figure 2* Bounded Box Wrapper with feature selection

For calculating landmarks, we have implemented Is frontal helper function that will return the landmarks if they are true and in between -25 and 25, if not then it will return false for every other value. Finally, for calculating the inference of the proposed model we placed time. Time () statements which returns time in seconds (float value) since the epochs.

### 4.1 Advantage of Proposed System

- The proposed system will provide faster facial detection and identification as compared to the previous one.
- The InsightFace module will provide lower time complexity while parallel increasing the accuracy of the Model
- The proposed system has increased dimension i.e., the system will scale as more users are registered with the system.

### 4.2 Proposed System Pseudocode

- Systems AI Model loaded:
    - Insight-face facial detection model
    - Insight-face facial recognition model
- Known faces images loaded and facial embedded extracted and saved
- Input feed captured through RGB camera
- Image frame passed through insight-face (arc face) facial. detection model
- Cropped face separated from each frame
- Cropped face passed through insight-face facial recognition model to extract embeddings

- Extracted embedded compared with known embeddings 8. Known face with the highest similarity identified as the person in frame
- Identified person emailed
- Desktop notification generated

### 4.3 Proposed System Design

When the program detects the user is not wearing the mask it first takes the input from the image, crop the image so that only the facial part will be available to send to the DNN model. After that first it represents the image by drawing Bounded Box Wrapper and then performs detection to identify the person and in the last it will notify the user by sending an e-mail and also it goes back to iterate through the face mask detection part again.

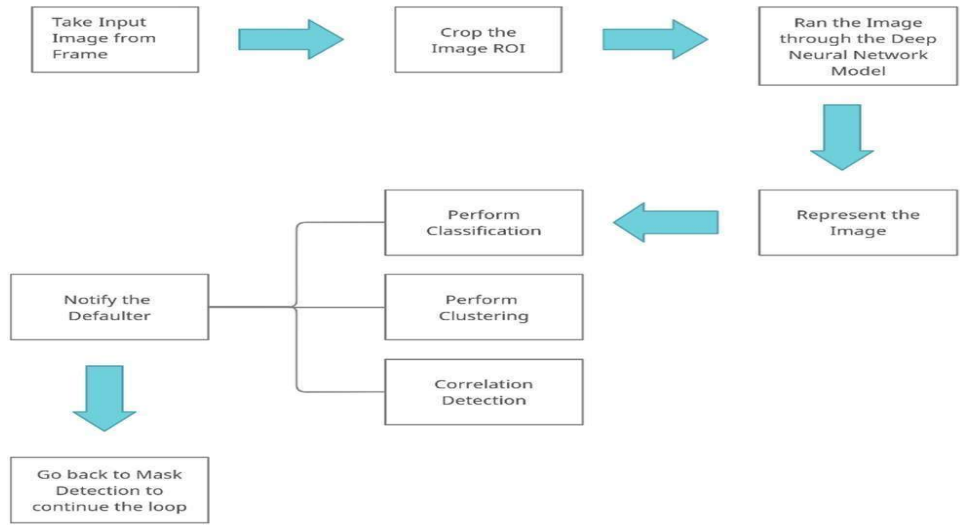

*Figure 3 Person Identification Model*

### 4.4 Complete System Design

As we have worked on the Person Detection and Identification part in this paper the other half detection Architecture will remain the same as defined in the previous paper and also shown in **Figure 3** Person Identification Model.

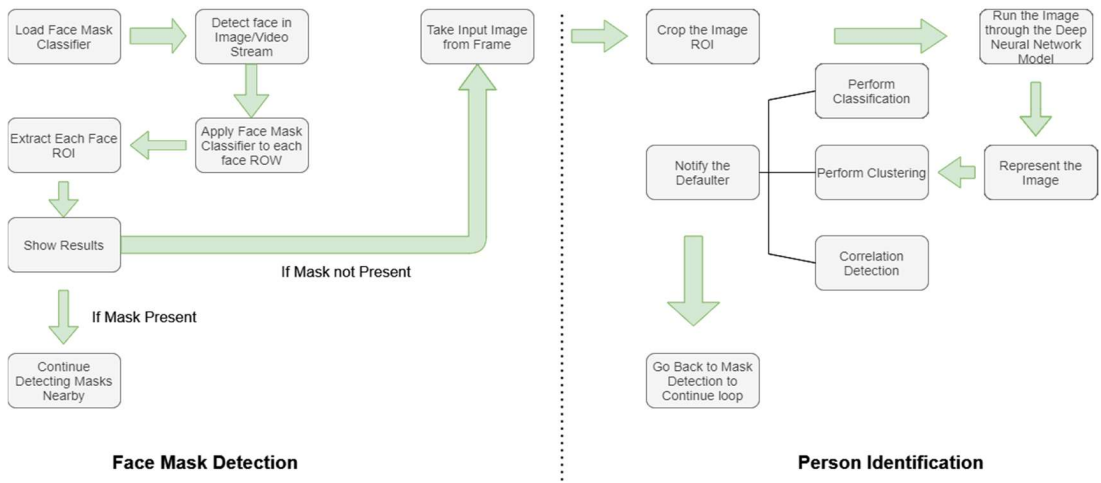

*Figure 4 Face Mask and Person Identification Module Workflow*

### 4.5 Formulating the Dataset

For creating the dataset for mask and no mask the data is collected from different free and open-source images websites Kaggle, google images, etc. The images are then classifying in multiple folders as masked and no mask. The first folder labelled as masked consist of 1915 entries whereas the second folder labelled as no mask consist of 1918 entries.

### 4.6 Training process:

1. After feature xi and weight W normalization, we get the cos θj (logit) for each class as (Wi)'xi.
2. We calculate the arccosθyi and get the angle between the feature xi and the ground truth weight Wyi.
3. We add an angular margin penalty m on the target (ground truth) angle θyi.
4. We calculate cos(θyi+m) and multiply all logits by the feature scales.
5. The logits then go through the SoftMax function and contribute to the cross-entropy loss.

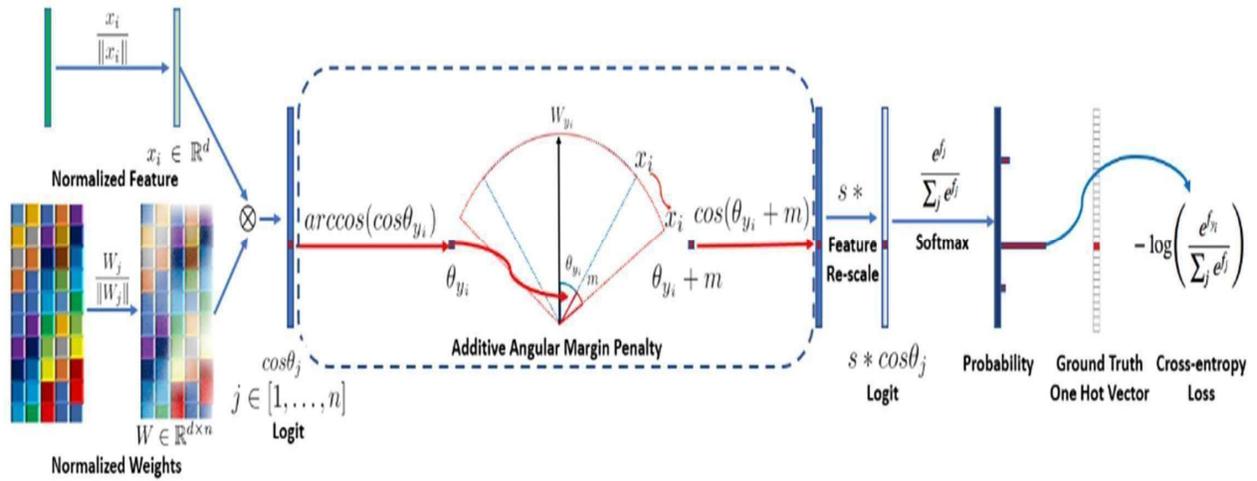

*Figure 5 Training Diagram*

## 5. Simulation Study

The section provides the benchmarking results of our proposed RFMPI-DNN model and the previous Detection model on WIDER FACE dataset.

### 5.1 Dataset

WIDER FACE is a largest public dataset available which contains 393,703 faces and 32,203 images in its database built on 61 event classes from internet [13]. The database consist of several humans faces, different poses, occlusion, expression, illumination and low-resolution. The images in the WIDER FACE dataset are divided into 3 classes for training, testing and validation purposes each of them is divided into a ratio of 50%, 10%, 40% respectively. Depending on the edge box each subset is defined into three difficulty levels that are "Easy", "Medium" and "Hard". The Hard level covers all the detections from easy and medium, meaning hard can show effectiveness of different methods. In our simulation, the proposed RFMPI-DNN model and the previous Detection model both uses WIDER FACE dataset containing 16,102 images from which 196,852 annotated faces are extracted [14].

### 5.2 Feature Extractor

We have used Resnet-10 as backbone and caffe-model as neck to construct the feature extractor [15]. The combination of both is used in most of detectors, so it would be good for comparison and replication.

Page **6** of **12**

## 5.3 Training Values and Test Size

We have trained both the model using the Stochastic Gradient Decent (SGD) optimizer (momentum=0.9, decay=0.01) having batch size =32 on GeForce GTX 1060ti [16]. The initial learning rate is 1 x e-4 as lower the initial learning rate the better the results. WE have used 40 EPOCS so that our training accuracy would be higher. EPOCS are hyper parameter i.e., it tells us about the number of times the training algorithm will work.

## 5.4 Comparison on WIDER FACE

The Table 1. Shows the Average Precision (AP) of our RFMPI-DNN model and the previous model on WIDER FACE test and validation subset. Our model performs better in all 3 segments of dataset NAD gets very promising results.

| METHOD | BACKBONE | EASY | VAL MEDIUM | HARD | EASY | TEST MEDIUM | HARD |
|---|---|---|---|---|---|---|---|
| PREVIOUS DETECTION MODEL | ResNet-10 | 0.969 | 0.958 | 0.921 | 0.965 | 0.957 | 0.9211 |
| RFMPI-DNN(OURS) | ResNet-10 | **0.972** | **0.965** | 0.925 | **0.967** | **0.962** | **0.924** |

Table 1 Average Precision Performance

Even when both the backbone is same in the model our model outperforms the previous model in each level in validation as well as in testing phase. The Precision Recall curves (AP) are shown in Fig 5. Our model achieves best AP in all level faces, i.e., AP = 0.972(Easy), 0.965 (Medium), 0.925 (Hard).

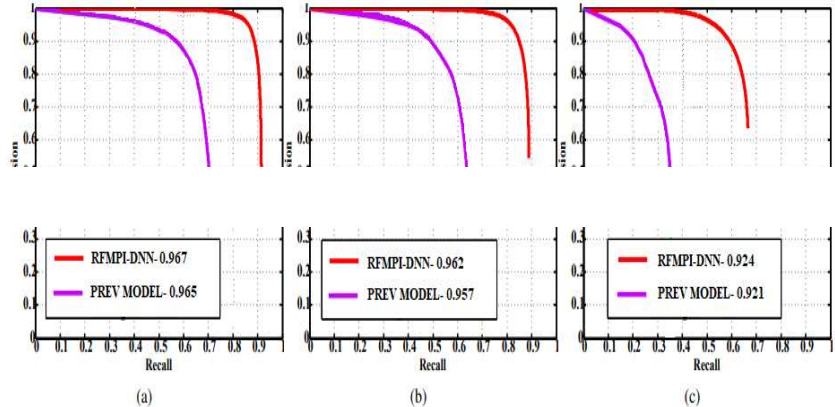

*Figure 6 Precision-Recall curves obtained by our proposed RFMPI-DNN red) and the previous detection model All methods trained and tested on the same training and testing set of the WIDER FACE dataset (a): Easy level, (b): Medium level and (c): Hard level.*

## 6. Result Discussion and Analysis

The graph is plotted in pythons matplotlib **Figure 6** Precision-Recall curves obtained by our proposed RFMPI-DNN red) and the previous detection model All methods trained and tested on the same training and testing set of the WIDER

FACE dataset (a): Easy level, (b): Medium level and (c): Hard level. Library which gives us the complete understanding why our RFMPI-DNN model is showing better result than the FACE RECOGNITION model used previously. The graph plots the time inference of four outputs i.e., Mask, No Mask, Face Detection and Person Identification. The time taken by both the model is showed as a comparison side by side.

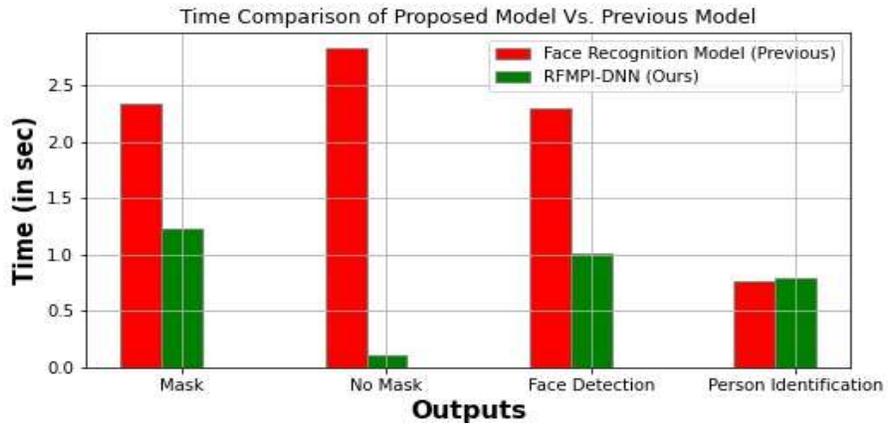

*Figure 7 Time Comparison of Previous and Proposed Model*

### 6.1 Connectivity

The connectivity between the two models will remain same as in previous paper i.e., The model first runs the face detection part to detect the face in the frame, if face not found it will continue until it founds one, when the face is found it will detect the person wearing a mask or not if the person is wearing a mask it will continue detecting other faces (if there are more faces available) or else if the no mask is detected the system will switch to person identification model the model fetched the face ROI from the image and compare it with the face data available in the database and gives out notification that the person is not wearing a mask. The Figure 7 Time Comparison of Previous and Proposed Model shows the diagram of connecting two models.

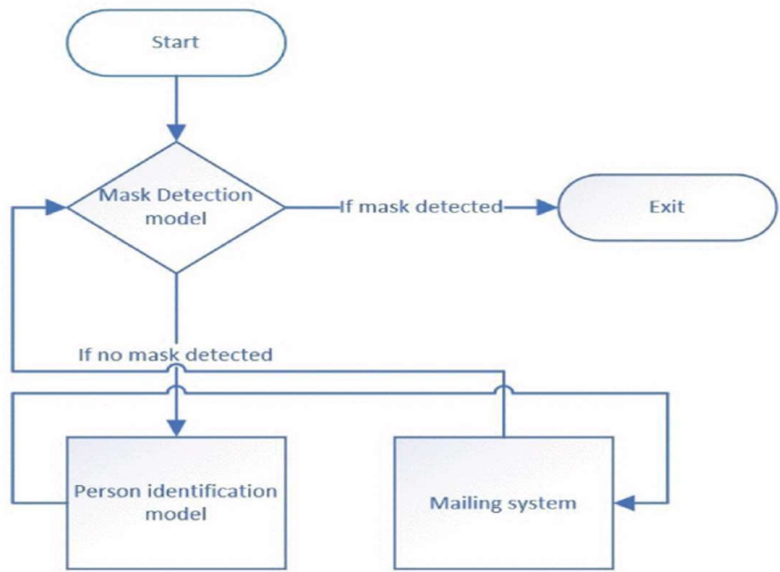

*Figure 8 Connection of Both Model*

### 6.2 Proposed Model Accuracy Samples

The proposed detection model will tightly bound the complete face while the previous model does not cover the complete face ROI Figure 8 Connection of Both Model. Shows the proposed model and previous model facial detection accuracy.

**Black Represents the Old Face Recognition Module**
**White Represents the New Face Recognition Module**

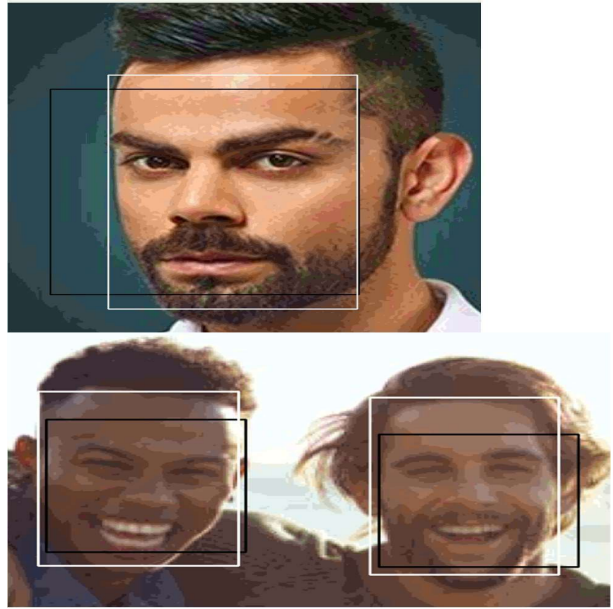

*Figure 9 Face Module Comparison*

White tightly covers the whole face while black is not tightly bound and does not cover the whole face.

### 6.3 Embedding Size

The old facial recognition model supports only supports dimension up to 128 means system does not have capability to detect more persons at a time. The proposed model has dimensions of 512 which means the system can support more people and will be reliable even if the number of people increases. Thus, the system will scale as more users are registered with the system. The same has been presented in a tabular form in **Error! Reference source not found.**.

|  | OLD DIMENSTIONS | NEW DIMENSIONS |
|---|---|---|
| EMBEDDING SIZE - FACIAL RECOGNITION | 128 | 512 |

*Table 2: Facial Recognition Dimensions*

### 5.3 Evaluation Results

The result of inference time a on Face Mask, No Mask, Face Detection and Person Identification presented in Fig 5.has been tabulated for better understanding of results in **Table 2**: Facial Recognition Dimensions. By looking at the table we can clearly say that our proposed model performed significantly better than the existing model.

| FUNCTIONS | OLD SYSTEM INFERENCE TIME | PROPOSED SYSTEM INFERENCE TIME |
|---|---|---|
| DETECT AND PREDICT MASK | 0.0234 | 0.0123 |
| DETECT AND PREDICT NO-MASK | 0.0283 | 0.0111 |
| FACE RECOGNITION | 0.230 | 0.0101 |
| PERSON IDENTIFICATION | 0.0077 | 0.0079 |

*Table 3: Inference Time Table*

### 5.4 Working of Model

**Figure 8** Connection of Both Model the model running and successfully detecting the person face mask and **Figure 10** shows the accuracy result around the edges of the box.

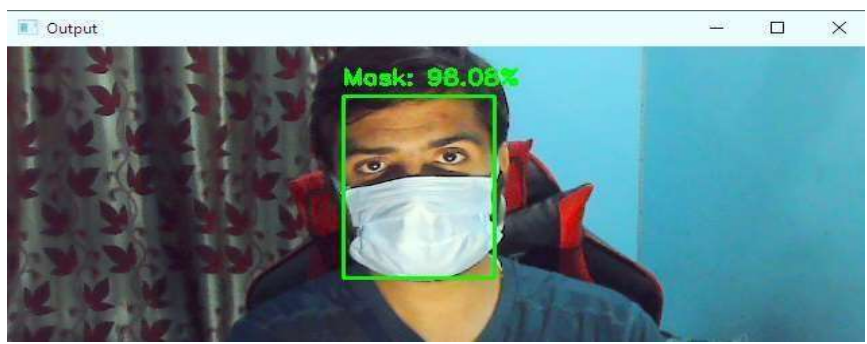

*Figure 10 Model Detecting Person wearing a mask*

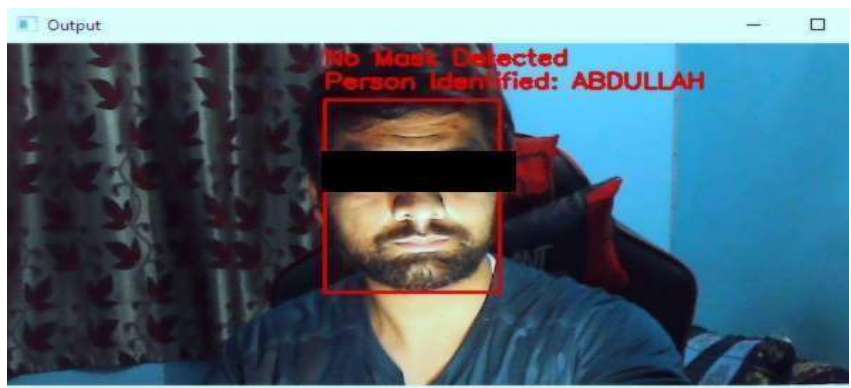

*Figure 11 Working of Person Identification Model*

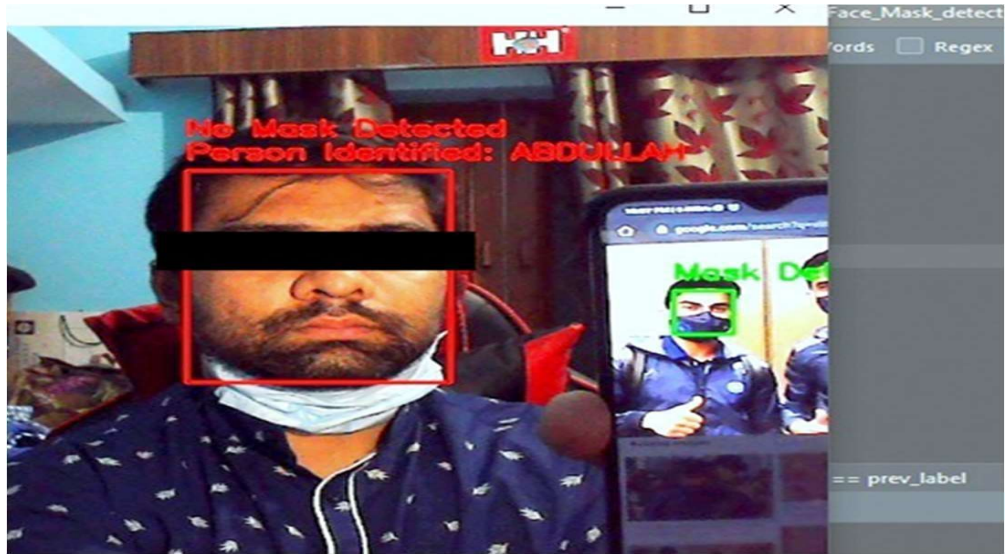

*Figure 12 Person ID and Mask Detection Model Combined*

**6.5 E-mail Notification**

Apart from getting notified on the screen the default individual which is not covering his/her face with a face mask will be notified by the system by sending an email to the person from the systems custom email id. This work is done using SMTP module of python.

## 7. Conclusion and Future Works

**7.1 Conclusion**

As the pandemic is still going on and currently there is no future prediction regarding when COVID-19 will successfully get over getting vaccinated is the option but wearing face mask is still a need even after getting vaccinated. This proposed model will deal efficiently even the larger groups of people and will detect and identify person more effectively and efficiently. Faster recognition is need of an hour because of how COVID-19 spreads. Apart from faster detection the user will get notification on his/her email address so that they would not forget to cover their face and stick to the terms set by the government for covering the face.

**7.2 Future Work**

Some of the future works can be considered are given below:
- Detection of Suspicious Person with covered face.
- Adding mobile message system for efficient and fast notification system.
- Adding alarm signal for alerting the person in the real time.
- Performing and adding new classification and detection model for getting better workflow of the system.